# Large Scale Diverse Combinatorial Optimization

ESPN Fantasy Football Player Trades

Aaron Baughman

IBM, USA, baaron@us.ibm.com

Daniel Bohm

Disney, USA, Daniel.bohm@disney.com

Micah Forster

IBM, USA, mforste@us.ibm.com

Eduardo Morales

IBM, USA, Eduardo.Morales@ibm.com

Jeff Powell

IBM, USA, jjpowell@us.ibm.com

Shaun McPartlin

Disney, USA, shaun.mcpartlin@disney.com

Raja Hebbar

IBM, USA, raja.hebbar@us.ibm.com

Kavitha Yogaraj

IBM, India, kyogarj1@in.ibm.com

Yoshika Chhabra

IBM, India, Yoshika.Chhabra@ibm.com

Sudeep Ghosh

IBM, India, Sudeep.Ghosh1@ibm.com

Rukhsan Ul Haq

IBM, India, rukhsan.ul.haq@ibm.com

Arjun Kashyap

IBM, India, Arjun.Kashyap@ibm.com

Even skilled fantasy football managers can be disappointed by their mid-season rosters as some players inevitably fall short of draft day expectations. Team managers can quickly discover that their team has a low score ceiling even if they start their best active players. A novel and diverse combinatorial optimization system proposes high volume and unique player trades between complementary teams to balance trade fairness. Several algorithms create the valuation of each fantasy football player with an ensemble of computing models such as Quantum Support Vector Classifier with Permutation Importance (QSVC-PI), Quantum Support Vector Classifier with Accumulated Local Effects (QSVC-ALE), Variational Quantum Circuit with Permutation Importance (VQC-PI), Hybrid Quantum Neural Network with Permutation Importance (HQNN-PI), eXtreme Gradient Boosting Classifier (XGB), and Subject Matter Expert (SME) rules. The valuation of each player is personalized based on league rules, roster, and selections. Similarly, the cost of trading away a player is related to a team's roster, such as the depth at a position, slot count, position importance, etc. Teams are paired together for trading based on a cosine dissimilarity score so that teams can offset their respective strengths and weaknesses. A knapsack 0-1 algorithm computes outgoing players for each team. Postprocessors apply analytics and deep learning models to measure 6 different objective measures about each trade, such as parity, pain, and fairness. Over the 2020 and 2021 National Football League (NFL) seasons, a group of 24 experts from IBM and ESPN evaluated trade quality through 10 Football Error Analysis Tool (FEAT) sessions. Our system started with 76.9% of high-quality trades and was deployed for the 2021 season with 97.3% of high-quality trades. To increase trade quantity, our quantum, classical, and rules-based computing have 100% trade uniqueness. This paper will discuss our diverse computing paradigms that value players, cost determinations, personalization experience, knapsack 0-1 algorithm and our experimental results. Throughout the 2020 season, we served over 239 million trade proposals and insights with over 55 million user interactions. In 2021, we introduced trade personalization so that the trade proposals were created around user preferences. We use Qiskit's quantum simulators throughout our work, which employ classical processors to emulate quantum hardware [qiskit.org].

## 1 INTRODUCTION

Football team owners start with a drafted roster that has a team performance ceiling. For example, a team with a low ceiling could fall short even in the rare circumstance when each player on the roster has an exceptional (85$^{th}$ percentile) performance. To improve a team's roster, managers can acquire players from the waiver wire or trade with other peer teams within their league. With the best players rostered through the draft, mutually beneficial trades can provide the most impact on a team's outlook.

Hundreds of combinations of teams and players produce a large decision space for team managers to find trades. To further increase the dimensionality of the search space, fantasy football managers can select personalization features that include a watchlist, player acquisitions, untradable selections, and position favorites. Overall, the total number of possible combinations when considering personalization is $4 * 10^{41}$. The overwhelming number of parameters causes many human-produced trades to be suboptimal.

Our novel system applies a diverse combinatorial optimization pipeline that combines classical and quantum computing to generate an average of 10 trades for each team owner. Each day throughout the 2021 season, we delivered over 2.5 million trade proposals for team managers to consider. Each trade balances egocentrism with the realism that both teams will only accept a trade if the expected value exceeds opportunity costs.

Ultimately, ESPN and IBM aimed to increase the engagement rate that measures the interaction of fantasy football users on the ESPN fantasy football mobile application. ESPN's 10 million fantasy users have spent a total of 1.8 billion minutes on ESPN's fantasy app, a 4% increase compared to last season. As a record high, fantasy football participation was up 21% when compared to the 2020 season. Participation is defined as an average of at least one hour spent per fan per week on the ESPN app.

## 2 ABBREVIATED RELATED WORKS

Trading players is the foundation of many fantasy sports games. The process enables two users to exchange one or more different fantasy players to improve their team's overall scores throughout the season. Other fantasy sports platforms have implemented their own unique way of assisting users in exchanging players within their league. The related works have not used Combinatorial Optimization to formulate such assistance. Our work proposes an original, diverse combinatorial optimization system that includes quantum and classical computing to suggest high volume and unique player trades, assisting users in making better trades within the ESPN Fantasy Football app.

## 2.1 Combinatorial Optimization

Combinatorial optimization problems are ubiquitous and difficult to solve. Many industries such as aerospace, transportation, logistics, economics, and software engineering utilize this technique. For example, existing works have been applied to a stochastic combinatorial optimization model for identifying the best coverage in software engineering unit testing [Wang]. Additional works include combinatorial optimization applying a machine learning approach to knapsack problems [Nomer]. Decision optimization has also been applied to satellite equipment layout to minimize offset, cross moments of inertia, and space debris impact risk [Yan'gang].

Within the field of quantum computing, hardware devices for these types of problems have recently been developed. For example, a multilevel solver using quantum processors has begun to emerge [Ushijuima-Mwesigwa]. The state of the art of combinatorial optimization problems in quantum computing has many solutions ranging from solving integer programming to hybrid quantum and classical variational algorithms. These algorithms can be implemented on near term nois quantum computers to find an optimal selection from a finite set of solutions. Quantum Approximate Optimization Algorithm (QAOA) is a quantum algorithm that attempts to solve one such combinatorial problem [Farhi]. Other works have applied quantum heuristics to limit the combinatorial search space [Amaro].

## 2.2 Optimization in Fantasy Sports

In the domain of fantasy sports, platforms such as fantasy football calculator and dynasty 101 provide insights into potential trade packages between fantasy team managers with applications such as Trade Calculator and Trade Analyzer [dynasty101.com, fantasyfootballcalculator.com]. These alternatives provide valuable insights for fantasy team managers in determining if a trade package is fair and valuable. Our approach differs from the existing prior work by utilizing 146 components when calculating player valuations. In addition, these systems can only apply insights to suggested trade packages that a user inputs. The previous works do not initiate or self-generate trade packages. Our trade assistant system combines combinatorial optimization techniques and state-of-the-art classical and quantum machine learning algorithms with subject matter expert rules to provide diverse and fair trade package suggestions to fantasy team managers.

Further, the current state of the art within decision optimization in fantasy sports utilizes machine learning to predict fantasy points and uses these predictions to optimize a team by selecting the best possible combination of players [Landers]. Their work applies a neural network to pick players and generate the winning team 58% of the time. Work by MH. A. A. Nomer depicts a less-performant but fully functional knapsack algorithm using neural networks to optimize a small number of predictors. Another work by Vladislav Haralampiev extends the analysis and provides a theoretical justification of the neural network approach for solving combinatorial optimization. Our work extends the state of the art by ensembling diversity within computing paradigms as an input into a general knapsack algorithm.

## 2.3 Feature Importance

Feature importance plays an essential role in the interpretability and explainability of a predictive model. Each feature understanding algorithm provides insight into the data, model, and dimensionality reduction that can improve the efficiency and effectiveness of a predictive model. Our system uses feature importance techniques to provide us with relative weights that contribute towards a player's valuation. First, Permutation Importance (PI) randomly shuffles data for each predictor to determine the decrease in model performance score. The decrease in the score becomes the predictor's importance [Konig]. The Accumulated Local Effects (ALE) algorithm was developed to help remove correlation effects between predictors while computing feature importance [Apley]. Several works attempt to measure feature importance within quantum computing [Chen, Otgonbaatar]. However, our PI and ALE approaches are implemented within classical computing to reduce the number of required qubits.

## 3 OVERALL ARCHITECTURE

The trade assistant system is a large-scale combinatorial and personalized trade recommender. Two continuously running batch applications and an offline ensemble of quantum models support the distributed trade assistant algorithms within the trade assistant bench application. The trade assistant algorithm engine runs three parallel broad player valuation and cost computations. The summaries about each player, such as boom and bust chance, sentiment, percentage-owned, valuation, opponent rank, games left, season actual, and season projection are saved into a relational database A second application, a trade content generator, queries the data about each player from a relational database. The rows are converted to JavaScript Object Notation (JSON) files organized by computing modes such as classical, SME, and quantum. Each JSON file is uploaded to Content Delivery Network's (CDN) origin, Cloud Object Storage (COS), for web acceleration and caching. The fast content access enables hundreds of parallel trade assistant algorithm bench applications to run and service requests concurrently.

The trade assistant algorithm bench has two purposes. First, it is consumed as a service by the Football Error Analysis Tool (FEAT) application. The low volume deployment captures user feedback for overall algorithmic training and evaluation before deployment to a production environment. User study groups and testers request trades with or without personalization selections from the bench application. For example, users can select their favorite players to add to a watch watchlist. Trades along with insights are returned and rendered on the FEAT application. Each user can rate each trade from the perspective of both trading partners. The ratings are used as labels for training and evaluation data. Second, the trade assistant algorithm bench was scaled to 525 Kubernetes pods in 2020 and 600 pods in 2021 across 3 OpenShift regional clusters for computing on demand. Hundreds of requests per second are routed to a cluster to create trades that consider league rules, rosters, teams, personalization selections, diverse computing paradigm metrics about each player, and trade filtering and evaluation criteria.

Fantasy football team managers can access the trade assistant through several entry points. A user can acquire or trade away a player directly from a player card within the ESPN mobile app. Alternatively, users can initiate trades through the ESPN Fantasy Football mobile menu system. Personalization selections can be completed through each user's profile or selections when creating a trade.

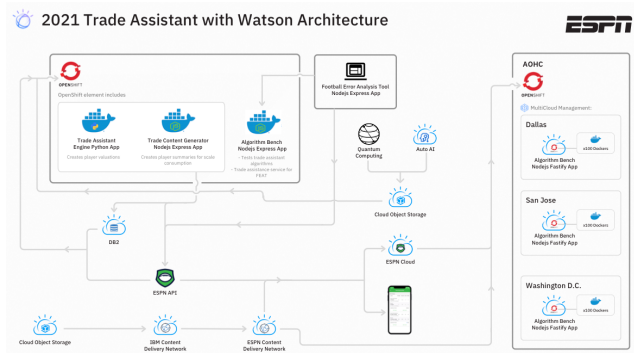

Figure 1: 2021 Trade Assistant Architecture

## 4 USER FOCUS GROUPS

An expert user focus group was selected to evaluate fantasy football trade recommendations to produce training data and evaluate the trades' overall quality. A total of 24 fantasy football experts from ESPN and IBM completed over 10 separate trade rating sessions with on average 500 ratings each before the 2020 and 2021 football seasons. Each user was assigned a league, team, and set of weeks to rank generated trades. Each participant rated every trade between 1 (lowest) and 10 (highest) for both teams' perspectives within FEAT. The rating from both sides of the trade was averaged together for a general assessment of the trade.

In 2020, the group was required to have at least 70% of trades rated higher or equal to 4 with 68% agreement, based on the kappa statistic. At the beginning of the ratings, the disagreement score was very high at just over 50%. The evaluators normalized on how to rate, such as minimizing the bias of future player outcomes to current trade suggestions. Within FEAT, additional context was provided around each player at the time of the trade, such as injuries, COVID-19 list, and suspensions. The full roster of each team was made available so that raters could view the strengths and weaknesses of each team at the time of trade generation.

The expert group and highest disagreeing pairs of raters met until the kappa score was at least 68%. Over time, the kappa score increased to 70%, such that disagreement about trade ratings was not an important factor when evaluating the quality of trades.

## 5 TRADE OPTIMIZATION

The exchange of assets between two teams should establish a fair market value for players, determine the value and cost of players with respect to the owning team, generate combinations of players around optimization criteria, evaluate the fairness and utility of a trade, and filter out low-quality trades. These steps will facilitate high-quality trades to the benefit of both trading teams. A high volume of unique trades is generated from diverse computing algorithms that include rules-based, classical machine learning, and quantum machine learning. Team managers can then view this large volume of diverse and high-quality trades.

The combinatorial problem of trade generation can be influenced by user selection. The valuation of trades is boosted by a team owner's positions of interest such as quarterback, targeted player acquisition, release of a player, player watchlist, and a list of players available for trade. The personalization of trades increases the relevancy of each proposed trade to a user.

### 5.1 Player Valuation

Each player within fantasy football is assigned a broad market valuation. The valuation provides a comparable player value of worth across all leagues. Three valuation pipelines are independently executed and run in parallel to determine diverse player valuations that create more trades than any single approach. In 2020, we started by creating a custom valuation set of rules and equations with input from ESPN experts called SME valuation and rules-based valuations. For the 2021 season, we added classical and quantum machine learning techniques to provide a market valuation for players.

The range of the player valuation for all computing types is the minimum, $sme_{low}$, and maximum, $sme_{high}$, of the Subject Matter Experts (SME) valuation of each day. The classical and quantum valuation ranges, $c_{t,high}$ and $c_{t,low}$ will determine a new normalized player valuation, $p_{i,t}$, of computing type, $t$, for a raw player valuation $x_{i,t}$. Each of the player valuations for a specific computing type will be comparable and can be subject to post-processing.

$$p_{i,t} = sme_{low} + \frac{sme_{high} - sme_{low}}{(c_{t,high} - c_{t,low})} * (x_{i,t} - c_{t,low}) \qquad (1)$$

The broad player valuations are changed to reflect league-specific situations. A cross valuation is determined to create an additional correlation between a pair of rosters that will be trading. A positional decay vector is created based on the depth of a position. For example, a player's value at a specific position within a trade is lower if the trading partner has much depth at that position. The valuation for a player will decrease based on the starter slot need. If a team must start 1 quarterback yet has 4 rostered, the player's value is reduced by 0.75. The normalized player valuation is related to both league rules and an opponent's roster.

*5.1.1 SME Valuation*

Within quantum computing simulation, we are limited by the number of available qubits and run time. As a result, groups of features about each player are prioritized into four tier ranks by ESPN experts to fit onto quantum computing simulations.

Within the first tier, a long-term outlook of the season-long projection for a player broken out by position, such as quarterback, tightend, and running back, provides a singular rank. Tier 2 rankings include boom ratio, bust ratio, and the next game projection. The boom ratio determines the fraction of games, **G**, the player, **pl**, scored over the 85th percentile of all players within the position over all weeks.

$$\frac{1}{G}\sum_{g=0}^{G} P_{boom}(pl)\,;\, P_{boom}(pl) = \begin{cases} 1: \text{Percentile}(actual_{pl}) \in [85,100] \\ 0 \end{cases} \quad (2)$$

The bust ratio determines the fraction of games a player scored below the 15th percentile of all players within their positions over all weeks.

$$\frac{1}{G}\sum_{g=0}^{G} P_{bust}(pl)\,;\, P_{bust}(pl) = \begin{cases} 1: \text{Percentile}(actual_{pl}) \in [0,15] \\ 0 \end{cases} \quad (3)$$

The second-tier groupings give us a ranking valuation associated with a short-term score projection for the next game within the context of historical boom and busts.

The third tier included features that ESPN experts ranked as the most important such as boom result from the current week, percentage of leagues that started a player, preseason projection, and season-long projection valuation. Shown in equation 4, the season-long projection valuation, PV, is the cumulative distribution function of the normal given ESPN's per position season-long points average, $\mu_{pts}$, standard deviation, $\sigma_{pts}$, and a player's projected season score, $x_{pts}$.

$$PV(x_{pts}, \mu_{pts}, \sigma_{pts}) = \int_{-\infty}^{x_{pts}} \frac{1}{\sqrt{2\pi}\sigma_{pts}} e^{\left[-\frac{(t-\mu_{pts})^2}{2\sigma_{pts}^2}\right]} dt,\, \sigma_{pts} > 0 \quad (4)$$

The final, fourth tier includes the player's average draft position taken over all leagues. The result provides a brand valuation for the player before the current league starts to play. Each of the ranking tiers is combined with expert opinions and experimental validations. A decay factor is added to the player brand to account for recency bias.

$$v_{sme,p} = \alpha_1 \text{tier\_rank}_1 + \alpha_2 \text{tier\_rank}_2 + \alpha_3 \text{tier\_rank}_3 + e^{(-week/3)} \text{tier\_rank}_4 \quad (5)$$

The rank for a player is penalized if the player is doubtful, inactive, out, probable, questionable, on the COVID list, or on injured reserve. An equivalence table normalizes positional values based on rank equivalence scores boosting, $\beta_1$, and expert opinions, $\beta_2$.

$$v\_norm_{sme,p} = \beta_1(pl_r, pl_p)\beta_2(pl_r, pl_p) \quad (6)$$

The valuation of a player considers performance momentum if a player has historical average points per game for the previous season. The $v\_norm_{sme,p}$ is weighted with a decayed rank contribution from last year's average points per game, $avg\_points_{prev}$. The scalers are determined by the week, *w*, which is the current scoring period that goes to 0 by week 6.

$$v\_final_{sme,p}(w) = \begin{cases} (6-w)v\_norm_{sme,p} + (w)avg\_points_{prev}: w \leq 6 \\ 6 * v\_norm_{sme,p} \end{cases} \quad (7)$$

The final valuations, $v\_final_{sme,p}(w)$, provide broad league-wide market estimations for each player.

### 5.1.2 Classical Valuation

Classical machine learning from Von Neumann computing provides traditional predictive modeling to discover the feature importance in trades. For each trade evaluated within FEAT sessions during the 2020 season, 146 predictors were extracted to describe the trade. A group of features described punditry opinion about a player that averaged media document sentiment around entity types such as game conditions, player performance, coach, team, location, health, and contract. State-specific predictors such as whether the player is a free agent, had a bye, injury status, is on injured reserve, had suspension, and is on bye provide estimations about individual players within trades. Descriptive statistics about the incoming and outgoing players within a trade, such as total incoming and outgoing counts of specific positions, provide important roster context. Performance predictors describing trades that include average low score, average high score, average projection, incoming season-long projection, low bust percentage, high boom percentage, and simulation projection quantify player game performance. Specific fantasy football statistics, such as percentage owned, percentage started, describe the trade within the context of fantasy sports.

A total of 4,733 exemplars were extracted from ESPN-rated trades. Trades were given an overall rating between a low score of 1 and a high score of 10. Any trade that was rated 4 or higher was accepted as a good trade. The label was reduced to a binary value that represented either a good or bad trade. The training data was balanced over both classes. An XGB Classifier outperformed decision tree, extra trees, gradient boosted tree, LGBM, logistic regression, random forest, and SVM models. The top 5 models are shown in table 1.

The XGB Classifier softmaxed feature importance provides us with the most important feature weights. The top 5 most important predictors for valuing a trade is shown in table 1. Features are extracted from a player and related to the trade predictors based on feature grouping. For example, the XGB weights about trade player ranks are combined into a single weight. The average document sentiment about incoming and outgoing players within a trade is combined into a weight about player document sentiment. The product sum of model predictor importance and player feature is penalized by a negative player state, such as is injured or suspended, with $p_{state}$. The final classical player valuation, $v\_final_{classical,p}$, is normalized to the same range as the SME valuation.

$$v\_final_{classical,p} = norm_{sme}\left(sme_{max}, sme_{min}, p_{state} \sum_{g=0}^{G} f_g w_g\right) \tag{8}$$

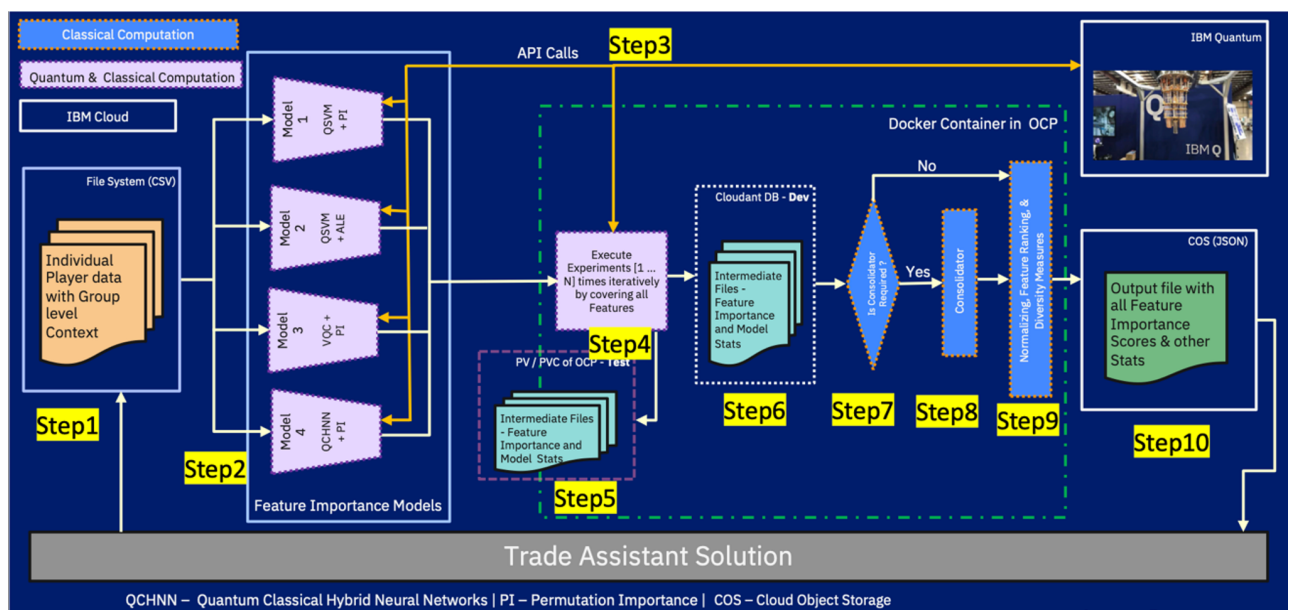

Figure 2: Quantum-Classical Computing Pipeline

*5.1.3 Quantum Valuation*

Quantum Machine Learning that leverages both quantum qubits and classical bits was used to value each player. A combination of quantum feature importance provides a diversity contribution to good trades. This results in the end-user seeing an increase in the volume of trades that balances fairness and value from each type of computing. The quantum algorithms are limited by the size, scale, and quantum systems currently available from IBM Quantum. The boundaries of the problem space are adjusted to conform to end-to-end qubit encoding size constraints while also finishing quickly enough to deploy to the ESPN Fantasy Football 2021 season. To emulate quantum systems, we used the Qiskit quantum simulators that match the quantum volume of available systems.

The overall quantum architecture is shown in Figure 2. The input file (step 1) is divided into tiers based on the input tiering strategy of grouping together features highly correlated to good trades. The current maximum capability we experimented with using a state vector simulator (as of 10/5/2021) is 21 features. We ran experiments for runtime analysis with 4,000 rows of data with 14 features that took 7 hours per tier. When we ran 4,000 rows with 21 features, the runtime took 22 hours per tier. We chose 14 features per tier to finish training in time to deploy the system for the 2021 season. Each tier contains 14 features and 4,000 rows such that each pipeline within step 2 of figure 2 has 56,000 exemplars for each of 11 tiers. In Step 4, all four feature importance models processed each of the tiers.

The objective of Quantum Machine Learning (QML) for the trade assistant was to process classical datasets with quantum algorithms trained with supervision [Havlicek]. A quantum SVM, the approach used in models 1 and 2 within Figure 2, uses a quantum kernel estimator to estimate the kernel function and optimize a classical SVM directly. The algorithm classifies a trade as good or bad, taking advantage of the large dimensionality of the quantum Hilbert space to obtain an enhanced solution. The Quantum kernel thus plays a vital role in the hybrid algorithm in separating the data with a hyperplane. We use the quantum computer to learn the kernel matrix for all pairs of points $\vec{x_i}, \vec{x_j} \in T$ in the train data.

$$K(\vec{x_i}, \vec{x_j}) = |<\Phi(\vec{x_i})|\Phi(\vec{x_j})>|^2 \qquad (9)$$

After choosing a feature map, we apply this quantum SVM classifier for the mapped data to create a hyperplane to separate good and bad trades.

The second quantum approach, model 3 in Figure 2, we implemented is called the Variational Quantum Classifier (VQC), which is helpful within the limits of current achievable quantum volumes. Classical data is mapped into a feature space with more dimensions than the classical space, making the problem of separating good and bad trades easier to solve. The VQC uses traditional variational methods with classical optimizers to train a parameterized quantum circuit. The classical data is encoded into a quantum embedding within a Hilbert space using a feature map. A combination of Hadamard and Pauli gates creates the encodings. A variational circuit is constructed using an ansatz, which exposes a set of variational parameters $\theta$ to perform the learning and tunning of this VQC model. VQC uses a classical optimizer to optimize a parameterized quantum circuit to provide a solution that cleanly separates the data.

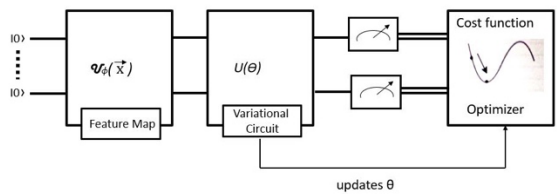

Figure 3: Variational Quantum Classifier circuit

Quantum Neural Networks (QNN) are used in model 4 of Figure 2 that have an edge over classical neural networks because, in QNNs, features are mapped onto a Hilbert space. While QNNs have an advantage in discovering small changes in the input features, they are computationally expensive and slow to train. To accelerate training, we combine classical artificial neural network models with quantum neural networks. QNNs are developed as feed-forward networks. Feed forward QNNs have two important calculations: forward propagation and backward propagation.

1. During forward propagation in the QNNs, the expectation value of the circuit is calculated. The output of a variational circuit (like the one used in QNNs) can be written as a quantum function $f(\theta)$, parameterized by $\theta = \theta_q, \theta_2, \ldots, \theta_N$ The partial derivative of this quantum function can be expressed as the linear combination of varied parameters passed to this quantum function. $\Delta\theta = f(\theta + 1) - f(\theta - 1)$

2. During backward propagation, the weights in a QNN are calculated using the parameter shift rule. The parameters of the quantum circuit are calculated with macroscopic shift. The gradient is then simply the output of the expectation value. Thus, we can systematically differentiate our quantum circuit as part of a larger backpropagation routine. This closed-form rule for calculating the gradient of quantum circuit parameters is the parameter shift rule**.**

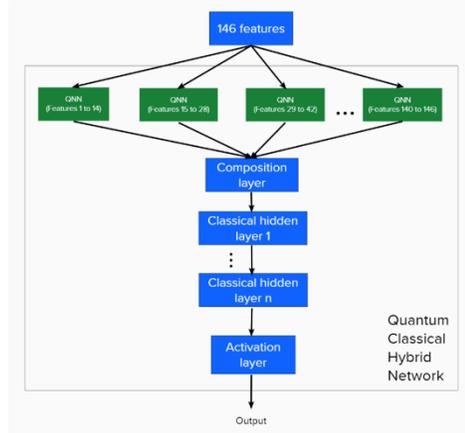

Figure 4: Quantum Classical Hybrid Network

In step 5, we interpret each model to extract the feature importance scores of their predictors. However, some machine learning models, such as for deep learning, are black box models. We applied Permutation Importance (PI, in Models 1, 3, and 4) and Accumulated Local Effects (ALE, in Model 2) for model interpretability and explainability. PI is a feature explainability technique that is independent of the model upon which it is executed. The permutation feature importance, $i_j$, is the decrease in a model score, **s**, when a single feature value is randomly shuffled over K iterations.

$$i_j = s - \frac{1}{K}\sum_{k=1}^{K} s_{k,j} \quad (10)$$

The ALE method determines feature importance based on a global explainability strategy. The entire data set is used to determine the average contribution of a given feature to a model's prediction accuracy. In ALE, we choose a local window for each feature and then find the effect on the model prediction as the window is traversed.

We combined our quantum algorithms with feature importance model interpretability methods. The four models within player valuation, as shown in figure 2, are:

1. QSVM with Permutation Importance
2. QSVM with ALE
3. VQC with Permutation Importance
4. Quantum Classical Hybrid Network

At the end of each parallel pipeline at steps 7 and 8, the tiered feature groups for a quantum model are combined based on model accuracy. Equation 11 shows how each feature's importance is reflected in the variable fir, based on a model grouping's accuracy.

$$\text{fir}_i = \frac{1}{2}(e^{\text{accuracy}_{k_i}} * x_i) + \frac{1}{2}(\tan(\text{accuracy}_{k_i}) * x_i) + x_i \quad (11)$$

Every feature, **i**, that is boosted by the tiered model accuracy and then normalized across all tiered models, j.

$$\text{fir\_norm}_i = \frac{\text{fir}_i}{\sum_{j=0}^{n} \text{fir}_j}; \sum_{j=0}^{n} \text{fir\_norm}_j = 1 \quad (12)$$

Each of the models' feature importance is combined and normalized.

*5.1.4 Player Valuation Diversity*

To compare quantum to classical computing, we measured diversity. To formalize our measurements, we introduced a notation called quantum diversity, $qd$.

$$qd = qa_k \mathbin{@} qrank\_diff_{avg} \mathbin{@} qvar \quad (13)$$

The first term, $qa_k$, is the quantum accuracy for model, k. The $qrank\_diff_{avg}$ is the diversity of feature rank, defined by the average percentage rank difference between the two pairings of quantum with classical, and quantum with SME.

$$q\_c_{\%rank\_diff} = \frac{\sum_{i=0}^{n} |q\_rank_i - c\_rank_i|}{max\_rank\_diff} \quad (14)$$

$$q\_sme_{\%rank\_diff} = \frac{\sum_{i=0}^{n} |q\_rank_i - sme\_rank_i|}{max\_rank\_diff} \quad (15)$$

$$qrank\_diff_{avg} = \tfrac{1}{2}(q\_c_{\%rank\_diff}) + \tfrac{1}{2}(q\_sme_{\%rank\_diff}) \quad (16)$$

The term $qvar$ is the average variance of the feature predictor importance over all quantum models. Over our computing paradigms, we calculate two summary diversity measures: quantum diversity, $qd$, and classical diversity, $cd$. For SME, we focus on accuracy since a predictive model was not used for ESPN feature rankings.

## 5.2 Player Cost

Every trade presents an opportunity cost for fantasy football player managers. The cost of trading away each player is related to the roster for a team, league rules, and performance estimations for a player. As such, the tradability cost is associated with each player that is rostered.

The tradability cost for a player is the average of the position importance, player score projection ratio to rostered players at that position, the ratio of player score projection to all rostered players, and the player positional rank.

$$pre\_pc_{i,t}\left(I_{t,i,pos}, R_{i,pos} Pl_{i,proj}\right) = \frac{1}{4}\left(\frac{I_{t,i,pos}}{max(I_{t,pos})} + \frac{Pl_{i,proj}}{\sum_{k=0}^{N} Pl_{k,proj}} + \frac{Pl_{i,proj}}{\sum_{j=0}^{position} Pl_{j,proj}} + \frac{R_{i,pos}}{max(R_{pos})}\right) \quad (17)$$

The positional importance for a player is the ratio of the maximum number of starters required to fill a slot to the number of players eligible for the slot. The $pre\_pc_{i,t}$ is normalized between $[0,1]$, which provides the $norm\_pcost_{i,t}$.

## 5.3 Team Pairing

Teams within fantasy football each have their own strengths and weaknesses that can often complement those of an opposing team for the benefit of both teams. Each team has certain positions such as tight end, quarterback, kicker, defense, wide receiver, and running back. Every position has an importance, $I_{t,p}$, and a strength, $S_{t,p}$, feature vector. The importance feature vector encodes the position, maximum slot number, and the number of players that can fill the slot. Flex positions are considered where multiple positions can fill a single slot. The strength feature vector represents the position, average position valuation, average position rank, average position projection, minimum position projection, maximum position projection, and the average position owned percentage. Importance and strength position vectors are concatenated together into a team vector, $\bar{t}_n$.

A pair of teams is selected by finding the most dissimilar teams that can complement each other. In the case of a candidate pairing, we calculate a dissimilarity score.

$$\theta_{n,n-1} = cos^{-1}\left(\frac{\bar{t}_n \cdot \bar{t}_{n-1}}{\|\bar{t}_n\| \; \|\bar{t}_{n-1}\|}\right) \quad (18)$$

The priority of team pairings is sorted in descending order from 90°, the most dissimilar. The top-ranked pair of teams is selected based on user input for trade package generation.

## 5.4 Trade Package Creation

The trading of players between teams is a significant combinatorial problem requiring continuous player valuation and cost. The problem is formulated into a 0-1 knapsack algorithm where each player can be selected at most once. The trade value is optimized under the constraint of a maximum cost related to the opponent's most valuable player and user risk selection.

To begin, each of the opposing team's player valuation and the current team's player cost is scaled by 100 and rounded to an integer.

$$\forall_i norm\_pval_{i,t} = ciel(pv_{i,t} * 100) \quad (19)$$
$$\forall_i norm\_pcost_{i,t} = ciel(pc_{i,t} * 100) \quad (20)$$

To create a trade, we maximize the normalized player valuation for a computing type, $t$, where each player can only be selected 0 or 1 times.

$$maximize \sum_{i=1}^{n} norm\_pval_{i,t} x_{ti}; x_{ti} \in \{0:1\} \quad (21)$$

The players selected for a trade will have an associated cost of releasing players. The overall cost should be less than the maximum cost the opponent is willing to accept. The maximum cost is equal to a selected risk, $\alpha$, multiplied by the highest cost of releasing a player on the opponent's roster, $C_{o,pmax}$. The risk is a selectable parameter on the ESPN fantasy football experience that scales the maximal player cost on the opponent's roster.

$$\sum_{i=1}^{n} norm\_pcost * x_{ti} \leq (\alpha * C_{o,pmax}); x_i \in \{0,1\} \quad (22)$$

The 0-1 knapsack algorithm is run two times for a pair of teams by swapping the current team and opponent team. The two generated lists of players to acquire are bundled into a trade package.

### 5.5 Trade Personalization

Fantasy football managers can select several attributes about players so that recommended trades are more relevant to their interests. Users can select a specific player they want to release or acquire. If a player is selected for inclusion, the 0-1 knapsack algorithm places that player with the candidate player list. When a player is to be excluded, the player is removed from the pool of available players.

Player valuations and costs can be altered based on user selections. A watch list of players that are important to a team owner is used to boost player valuations by weight, $w_1$, which increases the likelihood that a player from the set will be included within a trade. Owners can specify which players they prefer trading away. The player's tradability cost is reduced by $w_2$. A third weight, $w_3$, interprets a list of untradable players by increasing the cost of releasing a player to a percentile where the player will never be included within a trade. Finally, team owners can select football positions to target for trades. A position booster based on selected positions from defense, kicker, quarterback, tight end, wide receiver, and running back increases the player's value with $w_4$. The four weights, $w_1$, $w_2$, $w_3$, and $w_4$, were experimentally determined through FEAT sessions.

### 5.6 Trade Filtering

A series of filters and evaluation metrics provide a quantified assessment of each trade and help to reduce the tail of lower trade ratings. A trade parity score, $parity_{t1,t2}$, between two teams t1 and t2 determines the magnitudes of difference between a trade package's incoming and outgoing players.

$$parity_{t1,t2} = \frac{1}{2}\left(\left|\sum_{i=0}^{N} \frac{norm\_pval_{i,t1}}{\max(norm\_pval_{t1})} - \sum_{i=0}^{M} \frac{norm\_pval_{i,t2}}{\max(norm\_pval_{t2})}\right| + \left|\sum_{j=0}^{M} norm\_pcost_{j,t1} - \sum_{j=0}^{M} norm\_pcost_{j,t2}\right|\right) \quad (23)$$

An egocentric estimation that measures the trade impact, $impact_{t1}$, for a team owner compares the incoming player values to outgoing player costs. The metric quantifies the opportunity cost for the team manager.

$$impact_{t1} = \frac{\sum_{i=0}^{N} norm\_pval_{i,t1}}{\sum_{i=0}^{N} norm\_pcost_{i,t1}} \quad (24)$$

For a team manager to assess the value loss in a trade, equation 25 shows a summary of trade pain with respect to player costs and valuations.

$$pain_{t1} = \frac{\sum_{i=0}^{N} norm\_pcost_{j,t1}}{\sum_{i=0}^{N} \frac{norm\_pval_{i,t1}}{\max(norm\_pval_{t1})}} \quad (25)$$

We provide a trade upside prediction with an 11-layer deep neural network that considers both the utility and realism of a trade. Each trade package will have a cumulative upside that considers asset movement, risk, $parity_{t1,t2}$, $pain_{t1}$, and the similarity of the trading partners. The overall probability indicates a positivity metric that shows how helpful a trade is to all trading partners and the chance that both teams accept the trade.

A postprocessor applies 15 rule-based filters and 4 thresholds learned through FEAT sessions to remove the lowest quality trades based on human trade ratings. A thresholding method removes trades below an experimentally determined parity$_{t1,t2}$, pain$_{t1}$, differences between incoming and outgoing players. An upside prediction provides owners with a data point indicating if they should propose a trade to their opponent.

Over time, the rule-based filters were developed through consumer feedback through social media and discussions with the IBM and ESPN user group. For example, to keep teams competitive in the short and long term, any trades that did not allow a team owner to fill all starter slots are filtered. This extends to not trading away players if they are the only player within a position. Highly unrealistic trades such as swapping quarterbacks are removed. Further, simple player positional swaps that do not have personalization are not the most meaningful trades. We found that a best player parity rule was necessary not to trade a star players for many marginal players. If a kicker or defense is traded, we found that the best trades included another player. Only kickers and defense that are paired with another player are included within trades. To consider team positional depth, we filter out trades that include 3 or more players from the same position.

# 6 RESULTS

We assessed our system using both data science evaluation criteria and user group sessions. A new notation that includes accuracy and diversity measures provides a simple trade-off estimation. For example, the notation quantum diversity $qd_k$ relates quantum accuracy, $qa_k$, for model k to two diversity measures.

$$qd_k = qa_k @ qrank\_diff_{avg,k} @ qvar_k \tag{26}$$

The first diversity measure, qrank_diff$_{avg,k}$, averages model k's predictor importance percentage rank difference compared with both our manually labeled predictor importance by ESPN and classical model feature importance. The quantum rank difference provides a percentage ordinal rank difference measure.

$$qrank\_diff_{avg,k} = \frac{q\_c_{\%rank\_diff} + q\_sme_{\%rank\_diff}}{2} \tag{27}$$

The variance of the magnitudes of each predictor provides a second diversity measure.

$$qvar_k = \frac{1}{n}\sum_{i=0}^{P}(p_i - \mu)^2 \tag{28}$$

To compare quantum to classical, a classical diversity evaluation vector shows classical model accuracy, predictor percentage rank difference, and predictor magnitude variance.

$$cd_k = ca_k @ crank\_diff_{avg,k} @ cvar_k \tag{29}$$

After each of the classical and quantum machine learning models were evaluated, a total of 10 user group FEAT sessions were proctored so that a group of 24 human fantasy football experts could qualitatively rate suggested trades. The first user group study session consisted of validating the definitions of a good trade. Each evaluator had overlapping leagues, teams, and weeks to measure a disagreement score between the group. The group discussed ranking scales, what should be considered when rating trades, and how to provide qualitative feedback. At the end of the validation session, our kappa score reached 70%.

## 6.1 Football Error Analysis Tool

For the 2020 and 2021 ESPN Fantasy Football seasons, we collected labeled data from 10 Football Error Analysis Tool (FEAT) sessions. Each user had an account to log into FEAT. The user session administrator provided each user with a list of leagues, teams, and weeks to rate suggested trades. Every trade was rated twice from the perspective of each side of the trade. The ratings were between a low score of 1 where each trade did not provide any value to either team in the trade or a ten10 where each side's roster became significantly better. Both sides of the trade rating were averaged together for an overall trade rating.

The FEAT tool was flexible and modular, enabling different personalization modes and compute types to be hidden. For example, in the personalized selected player acquisition or release session, FEAT exposed the ability of users within the user group to select a player that must be included within a trade while all other features were hidden. Each of the ratings became our ground truth for model training and user evaluation.

After each session, post processors, the 0-1 knapsack, player valuation, player costs, or team pairings are adjusted. One of the most common qualitative feedback comments was that many of the trades were too complicated. We quickly eliminated any trade that had more than 3 players on the incoming or outgoing player list. When a trade was not personalized, we further simplified trades by filtering out trades that were not 1-1 or 2-2 player trades. For each change in the system, we ran subsequent FEAT sessions.

Even though we converged towards objective trade ratings, we had to consider user bias. Some users have favorite teams and players, which could affect trade rating. In each FEAT session, we had a few weeks of overlapping trades and a large selection of trade combinations for each user. This helped us to minimize *context-based bias*. Another source of user bias is *compute-based bias*. We did not want users to know which type of computing, quantum, classical, or rules based, was producing the trades. The compute type was blinded and shuffled for the user group. Instead of selecting a known compute type, each user could only pick A, B, and C.

As a result, the user sessions in FEAT maximized trade objectiveness while minimizing both context-based and compute-based bias. Overall, the user groups generated 4,732 labeled rows, where each label poses a binary classification example of either a good or bad trade.

### 6.2 Model Evaluation

Each of the models is trained with 3,786 exemplars and tested on 946 exemplars. As shown within table 1, the XGB Classifier was the top model within classical computing. Overall, it had the highest accuracy of 95.70%, slightly above the Hybrid QNN CNN-PI model. The Hybrid QNN CNN-PI model had the highest percentage rank difference when contrasted to classical and rules-based computing. The QSVM-PI, QSVM-ALE, VQC-IP, and Hybrid QNN CNN-PI had higher predictor magnitude variance, which is the second diversity metric.

Table 1: Trade Package model evaluation

| Model Name       | Accuracy | % Rank  | Predictor |
|------------------|----------|---------|-----------|
| XGB              | 95.70%   | 67.00%  | 0.002     |
| Hybrid QNN CNN-PI| 94.30%   | 79.96%  | 0.077     |
| QSVM-PI          | 85.50%   | 61.00%  | 0.069     |
| QSVM-ALE         | 85.50%   | 61.25%  | 0.003     |
| VQC-PI           | 57.3%    | 64%     | 0.003     |

Overall, the balance between accuracy and diversity was important when creating trades. The ensemble of quantum models weighted the predictor importance based on the relative accuracy, percent rank difference, and predictor variance between the quantum models. Each of the predictor weights is computed as follows:

$$\text{pred}_i = \sum_{m=0}^{M} \frac{\text{acc}_m * \text{pred}_i}{\sum_{n=0}^{N} \text{acc}_n} + \frac{\%\text{rank}_m * \text{pred}_i}{\sum_{n=0}^{N} \%\text{rank}_n} + \frac{\text{var}_m * \text{pred}_i}{\sum_{n=0}^{N} \text{var}_n} \tag{30}$$

Each of the predictors from every quantum model is weighted together. A player valuation is the product sum of all $\text{pred}_i$ weights and actual fantasy football player valuation.

### 6.3 Trade Ratings

A total of 24 fantasy football experts from ESPN and IBM completed over 10 separate trade rating sessions before the 2020 and 2021 football seasons. A good trade is recognized as a rating of 4 or higher. As shown in Figure 5 and at the start of the 2020 season, we had a large tail of bad trades rated 2 or lower. As a result, we deployed the trade assistant with a 76.9% ESPN accuracy. On the x-axis we have trade ratings from 1 to 10. On the y-axis we show the percentage of trades. By the end of the 2021 summer, we had almost completely removed the tail of bad trades. We changed how player valuation, player cost, and team pairing were done and added a few post processors to filter out complex trades. The entire curve from 2020 to 2021 shifted towards good trades.

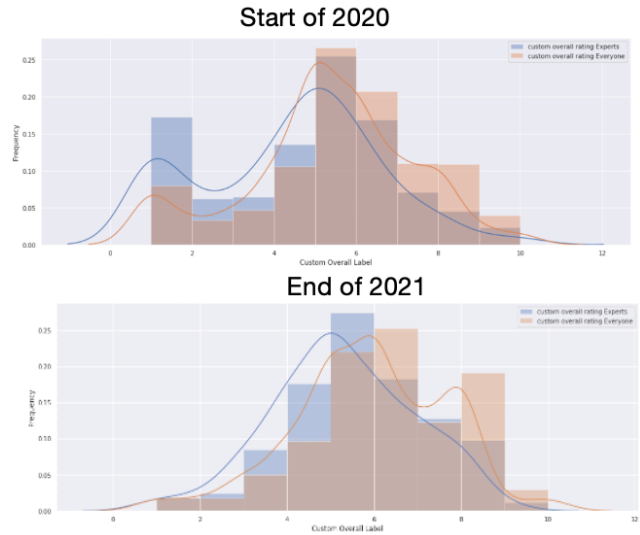

Figure 5: User group trade ratings

Each mode of computing improved upon the 2020 season's 76.9% trade accuracy. Quantum-classical computing trade valuation produced a 98.2% accuracy, followed by classical with 96.9% of trades rated as good. All three paradigms of computing produced excellent results and were deployed in 2021. The average rating for classical trades, 6.24, was the highest. Quantum was very close with 6.22, followed by rules-based (SME) with 6.11. The rating deviation was highest for classical.

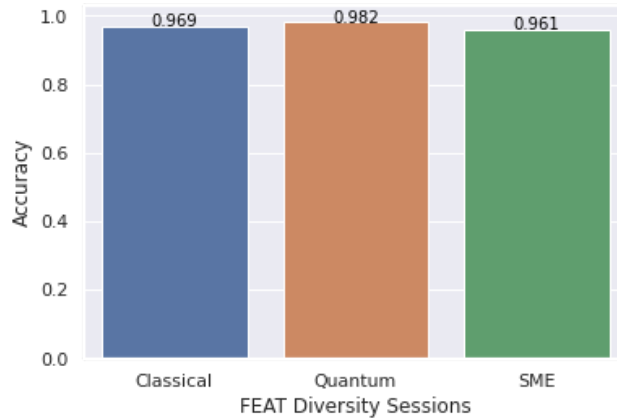

Figure 6: Computing paradigm trade accuracy

### 6.4 Trade Variety

The FEAT session was designed to measure the percentage of unique trades produced between quantum, classical, and rule-based computing. Every participant was given a unique combination of leagues, teams, and weeks. Every team pairing was ranked over all types of computing to compare sets of trades based on the computing paradigm. We had the following trade sample sizes 451, 307, and 292 for quantum, rules-based, and classical computing, respectively. Not every team combination will produce a trade. Across all trades viewed by fantasy football team managers, quantum-classical computing produced 60% of all trades, followed by 30% from rules-based and 10% from classical computing. We speculate that the ensembled QML approach contributed to more diverse trades. When we compared the trade sets of all users, every trade was unique across quantum, classical, and rules-based computing.

As a result, we produced highly accurate and diverse trades. In 2020, we averaged 3 to 4 trades per request. With quantum, classical, and rules-based computing combined, we averaged 10 unique and high-quality trades.

### 6.5 Trade Speed

The user experience on the ESPN Fantasy Football platform had to provide a response time of fewer than 5 seconds to maintain user engagement. In 2020, the trade assistant produced trades within 200 milliseconds with 200 requests per second load. The addition of quantum and classical computing with filtering and personalization took 1 second with 200 requests per second load. The inclusion of both high quality and diverse trades and the volume of trades added an acceptable 800 millisecond response time.

## 7 LARGE SCALE DEPLOYMENT

Throughout the 2020 season, we produced 2 million trade proposals with over 240 million insights about each trade. In 2021, we provided 2.5 million trades each day. With over 10 million users, we designed our system to compute broad player valuations and costs in continuous batch mode. A python application runs every 15 minutes to compute the top 400 fantasy football player classical, quantum, and rules-based player valuations. The predictive power for each variable in the quantum models and classical model is loaded through a CDN. We can run classical and quantum model builds offline to protect the load on our system. A NodeJS application converts the output of the player valuation and costs into JavaScript Simple Object Notation (JSON) files used by the consumer application.

In 2020, the trade assistant was scaled out onto 525 pods that run on RedHat OpenShift Kubernetes Service. For 2021, we increased the pod count to 600. Each quantum, classical, and rules-based player file is cached into each pod's memory to accelerate response time. A post request returns league rules, roster, and personalization requests to the trade assistant. All of the data is self-contained within each pod that produces trades. The algorithm pairs teams together for trades and modifies the player valuation and costs based on league rules and personalization requests. The precise fantasy football data tuned to a specific user is then input into the 0-1 knapsack algorithm. The trades are filtered for simplicity and quality before they are returned to the user.

## 8 FUTURE WORK

As the limits of quantum hardware are eased, we would like to run all our quantum machine learning models on quantum hardware. Today, we are running our workloads on a state vector quantum simulator. Even within the simulator, we are limited due to runtime. Running 21 features to train the model takes twice the computation time as training with 14 features. Within our application, the exponential increase in runtime relative to predictors is not practical. We want to train our quantum models on quantum hardware with a quantum volume to support 146 features. In addition, we would like to investigate the applicability of the Quantum Approximate Optimization Algorithm (QAOA) to gain a quantum operations research perspective.

From a user perspective, we will be experimenting with exposing quantum results within an understandable and transparent experience. We want to discern which trades are classical, quantum, and rules-based and how each helps the trading team. Contrastive and local explainability are topics that we will explore to produce a transparent understanding of each trade.

ACKNOWLEDGMENTS

We would like to thank ESPN, Michael Greenburg, Matthew Berry, Field Yates, Stephania Bell, Daniel Dopp, and Eli Manning for promoting and supporting our work. In addition, our gratitude goes to Noah Syken, Elizabeth O'Brien, John Kent, Tyler Sidell, Stephen Hammer, Gray Cannon, Chris Codella, Frederik Flother, and the IBM Quantum team.